\documentclass[journal]{IEEEtai}

\let\labelindent\relax
\usepackage{array}
\usepackage[caption=false,font=normalsize,labelfont=sf,textfont=sf]{subfig}
\usepackage{textcomp}
\usepackage{stfloats}
\usepackage{url}
\usepackage{verbatim}
\usepackage{booktabs} 
\usepackage{graphicx}
\usepackage{epstopdf}
\usepackage{amsmath,amssymb,amsfonts}
\usepackage{xcolor}
\usepackage{enumerate}
\usepackage{amsthm}
\usepackage{diagbox}
\usepackage{enumitem} 
\usepackage{color,colordvi}
\usepackage{epsfig}
\usepackage{mdwlist}
\usepackage{makecell}
\usepackage{multirow,multicol}
\usepackage{setspace}
\usepackage{float}
\usepackage{fancyhdr}
\usepackage[normalem]{ulem}
\usepackage{makecell}
\usepackage{fancyhdr}
\usepackage{flushend}
\usepackage{algorithm}
\usepackage{algorithmic}
\usepackage{cite}
\usepackage{multirow}
\usepackage{soul}
\usepackage[colorlinks,linkcolor=green,citecolor=green]{hyperref}
{}
{}
{}

\begin{document}

\title{Hierarchical Mutual Information Analysis: Towards Multi-view Clustering in The Wild} 
\author{{Jiatai Wang\href{https://orcid.org/0000-0001-7373-7706}{\includegraphics[scale=0.08]{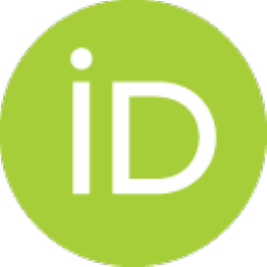}}}, 
        Zhiwei~Xu,~\IEEEmembership{Member,~IEEE,} 
        Xuewen Yang,~\IEEEmembership{Member,~IEEE,}
        Xin~Wang,~\IEEEmembership{Member,~IEEE,}
\thanks{This work  was supported by the National Science Foundation of China (61962045, 62062055, 61902382, 61972381),  the Science and Technology Planning Project of Inner Mongolia Autonomous Region (2019GG372).
(Corresponding Author: Zhiwei Xu.)
}
\IEEEcompsocitemizethanks{
\IEEEcompsocthanksitem Jiatai Wang is with the OPPO Research Institute, Beijing, China, 100020 (E-mail: 20211800675@imut.edu.cn).
\IEEEcompsocthanksitem  Zhiwei Xu is with Xinchuang Haihe laboratory, Tianjin, China, 300459, while visiting at Institute of Computing Technology, Chinese Academy of Sciences, Beijing, China, 100190 (E-mail: xuzhiwei2001@ict.ac.cn).\protect
\IEEEcompsocthanksitem Xuewen Yang is with InnoPeak Technology, Inc, CA, USA, 94303 (E-mail: xuewen.yang@protonmail.com).
\IEEEcompsocthanksitem Xin Wang is with the Department of Electrical and Computer Engineering, Stony Brook University, New York, U.S.A. 11794 (E-mail: x.wang@stonybrook.edu).
}
}	

\markboth{Journal of IEEE Transactions on Artificial Intelligence, Vol. 00, No. 0, Month 2023}
{First A. Author \MakeLowercase{\textit{et al.}}: Bare Demo of IEEEtai.cls for IEEE Journals of IEEE Transactions on Artificial Intelligence}

\maketitle

\begin{abstract}
Multi-view clustering (MVC) can explore common semantics from unsupervised views generated by different sources, and thus has been extensively used in applications of practical computer vision. Due to the spatio-temporal asynchronism, multi-view data often suffer from view missing and are unaligned in real-world applications, which makes it difficult to learn consistent representations. To address the above issues,  this work proposes a deep MVC framework where data recovery and alignment are fused in a hierarchically consistent way to maximize the mutual information among different views and ensure the consistency of their latent spaces. More specifically, we first leverage dual prediction to fill in missing views while achieving the instance-level alignment, and then take the contrastive reconstruction to achieve the class-level alignment. To the best of our knowledge, this could be the first successful attempt to handle the missing and unaligned data problem separately with different learning paradigms. Extensive experiments on public datasets demonstrate that our method significantly outperforms state-of-the-art methods on multi-view clustering even in the cases of view missing and unalignment.
\end{abstract}
\begin{IEEEImpStatement}
Ensuring the completeness and alignment of multi-view data is imperative for safe and stable operation of industrial processes, particularly under the case of suffering from data issues due to signaling during unsupervised learning. To eliminate the dependence on the assumption of data completeness in existing methods, this study simplifies complex model learning with instance-level and category-level strategies within a hierarchical mutual information framework. Our model is a multi-view extension of the general representation methods. It can further be applied to other applications, including but not limited to, cross-modal retrieval, autonomous driving, etc. By re-aligning and filling samples, our research can be used to explore consistency learning for each task, mining common semantics, and reducing decision risk. We thus suggest that people embed our model into the physical world to learn more consistent representation in broad scenarios and promote data-driven decision-making.
\end{IEEEImpStatement}

\begin{figure}[ht!]
\centering
\includegraphics[width=1\linewidth]{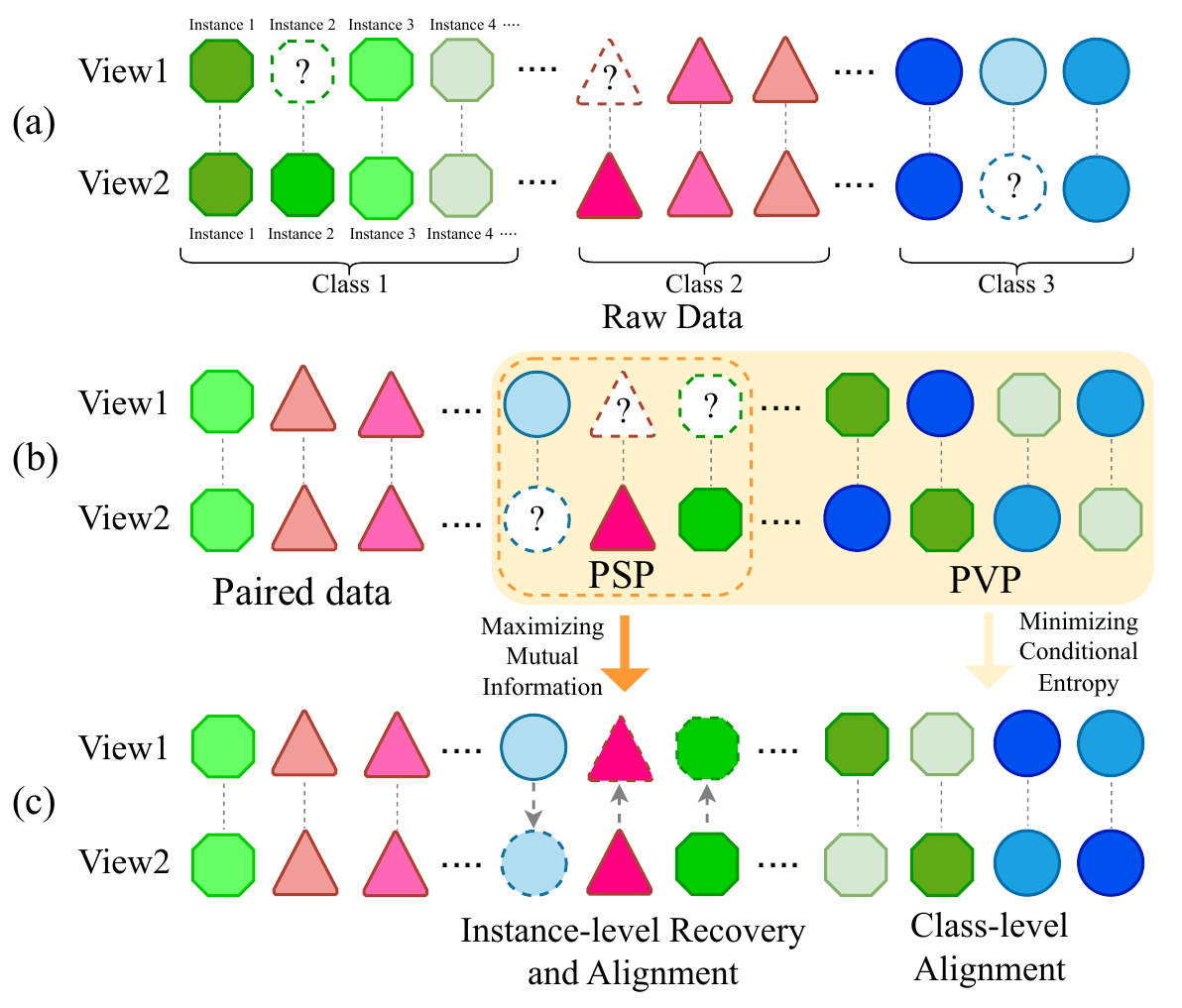}
\caption{Illustrative examples of the PVP and PSP. Taking a bi-view data as a showcase, we use two rows of polygons to denote two views, where each column of polygons represents a pair of instances that may be incomplete or unaligned as shown in (a). Polygons with the same shape belong to one category, and the same color is a pair of aligned instances. The "?" denotes that the view sample is missing. (a) Raw Data: there are missing and unaligned view samples in the raw data due to the complexity of data collection and transmission in practice. (b) Challenge: how to use paired data to recover missing instances and correct alignment to alleviate the negative effects of PVP and PSP. (c) Instance-level Recovery and Alignment: recovers each missing and unaligned views by using incomplete view samples of the same instance. Class-level alignment: Collect the samples belonging to the same class. }
\label{fig1}
\end{figure}
\section{Introduction}
\label{sec:intro}
\IEEEPARstart Data are often  collected from diverse sources in practical applications, and form multiple views. As an important unsupervised vision technology, multi-view clustering (MVC) aims to provide common semantics to improve the learning effectiveness and has made a remarkable progress. The success of all existing work \cite{MFLVC,deepMVC,msc,dmsc,SiMVC,IMVTSC-MVI,eamc}is supported under the assumptions of the completeness of data and the consistency of different views strictly. However, these two assumptions would be inevitably violated in real world (see Figure 1(a)). Firstly, Partially View-unaligned Problem (PVP) always exists due to the asynchronism of data transmission. Even worse, PVP even coexists with the Partially Sample-missing Problem (PSP) (see Figure 1(b)), while the data transmission of one view fails. It is a practical need and a challenge to learn common semantics in a robust way and alleviate the impacts of PVP and PSP to ensure the consistency of learning.

\begin{figure*}[h!]
  \begin{center}
  \includegraphics[width=1\textwidth]{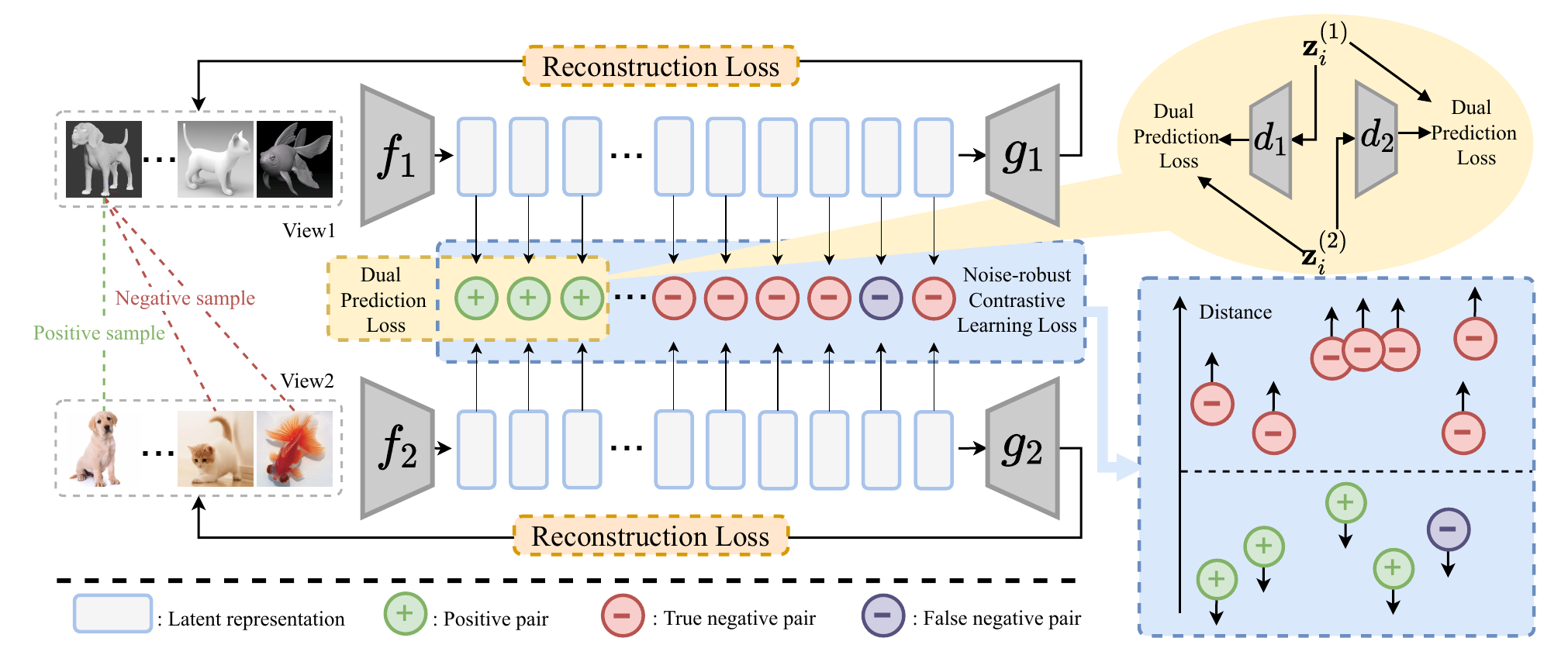}
  \caption{Overview of HmiMVC. In the figure, bi-view data is used as a showcase. As shown, our method contains three joint learning objectives, i.e., noise-robust contrastive learning, dual prediction, and reconstruction. To be specific, the noise-robust contrast learning achieves class-level alignment against PVP. Dual prediction allows constructing instance-level alignment as well as recovering its missing views from one of its existing views. The goal of reconstruction loss is to maintain the diversity of views to project all views into view-specific spaces.}
  \label{fig2}
  \end{center}
\end{figure*}

During past years, efforts have been devoted to solving the problem of incomplete data in MVC by imputing the missing and unaligned samples with various methods \cite{MvCLN,completer,DCP,AMVC,cdimc}. Previous MVC methods can be roughly classified into three subgroups, i.e., i) MVC methods for complete data \cite{DCCA,MFLVC,deepMVC,eamc,dmsc} strive to learn discriminative representations by utilizing the consistent and complementary information from different views; ii) PVP-oriented MVC methods \cite{MvCLN,PVC}, which build cross-view mappings at the instance level in an unsupervised manner; and iii) PSP-oriented MVC methods \cite{IMVTSC-MVI,completer,DCP,AMVC,cimic-gan,cdimc}, which utilize existing views by way of mathematical derivation or model prediction to recover lost views. Although important progress has been achieved by existing MVC methods \cite{completer,DCP,AMVC,cimic-gan,uimc,simc,cdimc,PVC,MvCLN} in solving PSP or PVP problem, few solve PSP and PVP jointly.
Besides the above methods, very recently, Yang et al. find that the correspondence of negative pairs in contrastive learning might be false, i.e., false negatives, and propose SURE \cite{SURE} to handle that.
SURE imputes the missing sample by the weighted sum of its peers in the same view, and then re-aligns it by identifying same-class samples through contrastive learning, which establishes class-level correspondences to unify the handling of PVP and PSP. However, it brings two major drawbacks: i) solving both PSP and PVP makes the algorithm untargeted, and ii) the class-level cannot provide detailed information for each sample in the dataset compared to the instance-level leading to lower granularity and accuracy.

Actually, the different instances are also relevant in real world, which is more valuable for the solution of PVP.  
On the other hand, consistency between views within instances is key to PSP \cite{completer}. Correspondingly, the solutions for PSP and PVP should be scheduled on different levels.
In this paper, we study this hierarchical relation between PVP solutions, those for PSP within a hierarchical mutual information framework as well as  an unified consistent framework to solve both PSP and PVP problems.
The proposed model is designed by using distinct scales of mutual information since missing and unaligned data exist in different hierarchies.
From the perspective of information theory, mutual information and the conditional entropy between different views measure the consistent and inconsistent semantics of multi-view data, respectively \cite{completer,DCP}. In addition, we introduce instance-level filling to further enhance the performance of clustering. Thus, we aim to maximize mutual information for PVP while minimizing instance-level conditional entropy for PSP, so that PVP and PSP are hierarchically handled while mutually enhancing each other.

In this work, we propose a novel incomplete multi-view clustering method, termed multi-view clustering via maximizing hierarchical mutual information (HmiMVC), to conquer the dual challenges of PVP and PSP caused by the lack of correspondence between views.
HmiMVC projects a raw dataset into a hierarchical latent space wherein information consistency is guaranteed. As shown in Fig. \ref{fig1}(c), we first solve PVP by maximizing mutual information through the contrastive learning to achieve the class-level alignment. To solve  both the problems of PSP and PVP, we introduce the idea of pairwise prediction to predict the missing data while minimizing the conditional entropy, which constitutes a natural instance-level alignment. Finally, we merge two levels of alignment strategies into a unified reconstruction process in a hierarchically consistent way to avoid the model collapse. The main contributions of this paper are: 
\begin{itemize}
    \item  we propose a novel integration of class-level and instance-level techniques to simultaneously address both the problems of PVP and PSP.

    \item From an information-theoretic insight, the proposed HmiMVC method is with a novel loss function which achieves the information consistency and data restorability using a contrastive loss and a dual prediction loss.
    
    \item Extensive experiments demonstrate the effectiveness and efficiency of HmiMVC in boosting mutual information, and state-of-the-art clustering performance and robustness.
\end{itemize}
\section{RELATED WORK}
\label{sec:RELATED WORK}
In this section, we briefly review two lines of related work, multi-view clustering and contrastive learning.
\subsection {Multi-view Clustering}
PSP-oriented MVC methods utilize the information contained in the existing cross-view counterparts to recover missing samples, which can be roughly divided into four categories: MVC based on kernel learning \cite{47,60}, MVC based on matrix factorization, MVC based on graph, and MVC based on spectral clustering.
Trivedi et al. proposed a kernel-based method \cite{60} that recovers a kernel matrix of an incomplete view from the kernel matrix of complete views, but it requires one of views to be complete. To address this limitation, matrix factorization-based MVC methods \cite{3, 4, 5, 46, 49} use lower ranks to project parts of the data into a common subspace, similar to the K-means relaxation method \cite{12}. However, feature-based matrix factorization methods generally cannot explore nonlinear data structures. This limitation is addressed by graph-based methods \cite{MNIST-USPS, cdimc, 64}, and methods based on  spectral clustering\cite{simc,62,65}. 
In recent years, deep learning-based multi-view clustering methods \cite{DECAF,dmsc,26,MFLVC,DCP,MvCLN,SURE} have received more and more attention. They exploit the excellent representation through the latent clustering patterns extracted from multi-view data. Researchers have made an excellent progress on PSP using a variety of methods \cite{completer,DCP,AMVC}, PVP is a less-touched problem revealed in very recent \cite{PVC}. However, PVP can only find the optimal alignment path using the traditional Hungarian algorithm \cite{hungarian}. Subsequently, PVC \cite{PVC} has implemented differentiable Hungarian algorithms, and MvCLN \cite{MvCLN} has achieved the class-level alignment using the robust contrast loss. Although SURE \cite{SURE} has extended the filling of missing views based on MvCLN, its class-level annotations cannot provide detailed information about the data samples, which makes it difficult for the model to distinguish between samples that are similar but belong to different classes resulting in information loss.

The differences between existing approaches and our HmiMVC are two-fold. On the one hand, HmiMVC is the first to use two different learning paradigms to separately handle PVP and PSP, and these two paradigms are implemented within the same information-theoretic framework to mutually boost each other. On the other hand, HmiMVC incorporates a fusion of class-level and instance-level approaches, where the instance-level granularity facilitates the algorithm in capturing finer details and features. This combination enhances the model's capacity to learn more intricate patterns and discriminative information from the data.

\subsection {Contrastive Learning}
Contrastive learning \cite{smiclr,moco,byol,simamse,dino} is an essential method for unsupervised learning \cite{31}. Its major goal is to maximize  the feature space similarity between positive samples while increasing the distance between negative samples. In the field of computer vision, contrastive learning methods have produced excellent results~\cite{32}. For example, SimClR\cite{smiclr} or MoCo\cite{moco}  minimize the InfoNCE loss function \cite{34} to maximize the lower bound of mutual information. Since the processing of negative samples is very cumbersome, the follow-up work, BYOL \cite{byol}, SimSiam \cite{simamse}, and DINO \cite{dino} have successfully transformed the contrastive task into a prediction task without defining negative samples and achieved amazing results. Therefore, contrastive learning with various objectives that maximize mutual information can help enhance the model's generalization capability and performance.

Previous work simply constructed positive and negative samples based on data augmentation. Although existing studies \cite{MFLVC,moco,iic} have shown that the consistency could be learned by maximizing the mutual information of different views, they ignore the mutual information at different hierarchies.
Lin et al. demonstrated that inconsistency in learning can be defined in terms of conditional entropy. Consequently, our strategy of learning mutual information from other hierarchies can be seen as an effective means to alleviate inconsistent learning \cite{completer, DCP}.
In contrast, HmiMVC not only uses the robust contrastive learning to reduce the impact of false-positive samples to solve PVP but also minimizes the conditional entropy to cope with PSP. Additionally, our method is specifically designed for handling missing 
unaligneddata, whereas the existing contrastive learning works ignore this practical problem.

\section{Method}
\label{sec:HmiMVC}
In this section, we propose a new deep multi-view clustering method, HmiMVC, to learn the representation of incomplete and unaligned multi-view samples in different hierarchies. As illustrated in Fig.\ref{fig2}, HmiMVC consists of three learning objectives: 
\begin{equation}
\label{eq1}
\begin{aligned}
\mathcal{L}=\mathcal{L}_{c l}+ \mathcal{L}_{\text {pre }}+\mathcal{L}_{\text {rec }},
\end{aligned}
\end{equation}
where $\mathcal{L}_{c l}$, $\mathcal{L}_{pre}$, and $\mathcal{L}_{rec}$ are noise-robust contrastive loss, dual prediction loss, and view reconstruction loss, respectively. For clarity, we will first introduce the proposed loss function and then elaborate on each objective.

\subsection{Notations}
A multi-view dataset $\overline{\mathbf{X}}_{N}=\left\{\mathbf{X}_{N_{x}}^{(v)},\mathbf{S}_{N_{s}}^{(v)},\mathbf{W}_{N_{w}}^{(v)}\right\}_{v=1}^{V}$ includes $N$ samples across $V$ views,
where $v \in[1, V]$ denotes the view index. 
$\left\{\mathbf{X}_{N_{x}}^{(v)}\right\}_{v=1}^{V}=\left\{\mathbf{x}_{1}^{(v)}, \mathbf{x}_{2}^{(v)}, \ldots, \mathbf{x}_{N_{x}}^{(v)}\right\}_{v=1}^{V}$ denotes the complete alignment data used for training, where $N_{x}$ is the number of complete and aligned instances.
$\left\{\mathbf{S}_{N_{s}}^{(v)}\right\}_{v=1}^{V}$/$\left\{\mathbf{W}_{N_{w}}^{(v)}\right\}_{v=1}^{V}$ denotes the data with  PVP/PSP, where $N_{s}$ is the number of unaligned instances and $N_{w}$ is the the number of missing instances( $N = N_{x} + N_{s} +N_{w}$).
The ratios of the set with missing data $\alpha$, the set with unaligned data $\beta$, and the set with complete data $\gamma$ are equal to $\frac{N_{w}}{N}$, $\frac{N_{s}}{N}$, and $\frac{N_{x}}{N}$, respectively.

\subsection{Class-level Alignment.}
Consistency learning can used to ensure the alignment of different views \cite{PVC}, and we exploit the contrastive learning to improve the learning of mutual information between different views. As shown in Fig. 2, we use $\left(\mathbf{x}_{i}^{(1)}, \mathbf{x}_{i}^{(2)}\right)$ as positive pairs ($i<N_{x}$), and stochastically select cross-view samples to form negative pairs $\left(\mathbf{x}_{i}^{(1)}, \mathbf{x}_{j}^{(2)}\right)$. Considering the impact of false-negative samples \cite{SURE}, mathematically,
\begin{equation}
\label{eq2}
\begin{aligned}
\mathcal{L}_{c l}=\frac{1}{2 N} \sum_{i=1}^{N}\left(Y \mathcal{L}_{i}^{p o s}+(1-Y) \mathcal{L}_{i}^{n e g}\right),
\end{aligned}
\end{equation}
where $N$ represents the total number of sample pairs, and $Y = 1/0$ for positive/negative pairs. Then the feature contrastive loss between $\mathbf{x}_{i}^{(1)}$ and $\mathbf{x}_{i}^{(2)}$ for the $i$-th pair of positive samples is formulated as:
\begin{equation}
\label{eq3}
\begin{aligned}
\mathcal{L}_{i}^{\text {pos }}=\left\|f_{1}\left(\mathbf{x}_{i}^{(1)}\right)-f{2}\left(\mathbf{x}_{i}^{(2)}\right)\right\|_{2}^{2}=\left\|\mathbf{z}_{i}^{(1)}-\mathbf{z}_{i}^{(2)}\right\|_{2}^{2},
\end{aligned}
\end{equation}
where $f_{1}$ and $\mathbf{z}_{i}^{(v)}$ denote the encoder and the latent representation of $\mathbf{x}_{i}^{(v)}$, respectively. We aim to maximize $\left(\mathbf{z}_{i}^{(1)}, \mathbf{z}_{j}^{(2)}\right)$'s distance in a latent space by minimizing 
\begin{equation}
\label{eq4}
\begin{aligned}
\mathcal{L}_{i}^{n e g}=\frac{1}{\tau} \max \left(\tau \left\|\mathbf{z}_{i}^{(1)}-\mathbf{z}_{j}^{(2)}\right\|_{2}^{{\frac{1}{2}}}-\left\|\mathbf{z}_{i}^{(1)}-\mathbf{z}_{j}^{(2)}\right\|_{2}^{{\frac{3}{2}}}, 0\right)^{2},
\end{aligned}
\end{equation}
where $\tau$ is the temperature parameter computed only once at the initial state with $\tau=\frac{1}{N_{p}} \sum d\left(\mathbf{x}_{i}^{(1)}, \mathbf{x}_{i}^{(2)}\right)+\frac{1}{N_{n}} \sum d\left(\mathbf{x}_{i}^{(1)}, \mathbf{x}_{j}^{(2)}\right)$, $N_{p}$ and $N_{n}$ denote the number of positive and negative pairs, respectively. In the inference phase we have $\sum_{v_{1}}^{V} \sum_{v_{2} \neq v_{1}}^{V} C\left(\mathbf{s}_{i}^{\left(v{1}\right)}, \mathbf{s}_{j}^{\left(v{2}\right)}\right)=V(V-1)$, realigning  $\mathbf{s}_{i}^{\left(v{1}\right)}$ and $\mathbf{s}_{j}^{\left(v{2}\right)}$ by computing the Euclidean distance $C(.)$.

\subsection{Instance-level Recovery and Alignment.}
To solve PSP in PVP, we train a group of dual prediction networks to minimize the conditional entropy \cite{DCP}. This network not only fills in the missing data but also achieves the natural instance-level alignment. Based on the definition of  Normalized Mutual Information, $\operatorname{NMI}(\mathbf{z}_{i}^{(1)}, \mathbf{z}_{i}^{(2)})=\frac{H(\mathbf{z}_{i}^{(1)})-H\left(\mathbf{z}_{i}^{(1)} \mid \mathbf{z}_{i}^{(2)}\right)}{H(\mathbf{z}_{i}^{(1)})+H(\mathbf{z}_{i}^{(2)})}$, minimizing the conditional entropy $H\left(\mathbf{z}_{i}^{(1)} \mid \mathbf{z}_{i}^{(2)}\right)$ maximizes NMI. According to the variational inference \cite{completer}, we introduce a network $d_{(v)}$ that minimizes the conditional entropy approximately by minimizing $\mathbb{E}_{\mathcal{P}_{\mathbf{z}^{(1)}, \mathbf{z}^{(2)}}}\left\|\mathbf{z}_{i}^{(1)}-d^{(2)}\left(\mathbf{z}_{i}^{(2)}\right)\right\|_{2}^{2}$. Further we have

\begin{equation}
\label{eq5}
\begin{aligned}
\mathcal{L}_{\text {pre }}=\left\|d^{(1)}\left(\mathbf{z}_{i}^{(1)}\right)-\mathbf{z}_{i}^{(2)}\right\|_{2}^{2}+\left\|d^{(2)}\left(\mathbf{z}_{i}^{(2)}\right)-\mathbf{z}_{i}^{(1)}\right\|_{2}^{2}.
\end{aligned}
\end{equation}
After the model converges, we predict the missing view and form a natural instance-level alignment, $i.e.$, ${\mathbf{w}}^{(1)}_{i}=d_{2}\left(f_{(2)}\left({\mathbf{w}}^{(2)}_{i}\right)\right)$, where $\mathbf{w}^{(1)}_{i}$ is the missing sample predicted to be recovered by the representation of $\mathbf{w}^{(2)}_{i}$. So $\mathbf{w}^{(1)}_{i}$ and $\mathbf{w}^{(2)}_{i}$ are aligned on the instance-level.

\subsection{Reconstruction for Hierarchical Consistencies}
 For each view, we pass it through an autoencoder to learn the latent representation $\mathbf{z}^{(v)}$ by minimizing
\begin{equation}
\label{eq6}
\begin{aligned}
\mathcal{L}_{rec}=\frac{1}{2 N_{x}} \sum_{i=1}^{N} \sum_{v=1}^{2}\left\|\mathbf{x}_{i}^{(v)}-g_{v}\left(\left[\mathbf{z}_{i}^{(1)}, \mathbf{z}_{i}^{(2)}\right]\right)\right\|_{2}^{2}
\end{aligned}
\end{equation}
As a result, the conflict between the reconstruction objective and two consistency objectives is alleviated, and trivial solutions are avoided.

\section{EXPERIMENTS}
\label{EXPERIMENTS}
In this section, we evaluate the proposed HmiMVC method on four widely-used multi-view datasets and compare it with three state-of-the-art clustering methods.

\subsection{Data description and Experiment setting}
Four widely-used datasets are used in our experiments. 1) \textbf{Scene-15} \cite{Scene-15} consists of 4,485 images distributed over 15 Scene categories, and we use two views of PHOG \cite{PHOG} and GIST \cite{GIST} features, 20D and 59D feature vectors, respectively. 2) \textbf{Deep Animal} consists of 10,158 images from 50 classes, and includes two types of 4096-dim features of \cite{Animal} extracted by DECAF \cite{DECAF} and VGG19 \cite{VGG19} respectively as two views. 3) \textbf{MNIST-USPS} \cite{MNIST-USPS} is a popular handwritten digit dataset, which contains 5,000 samples with two different styles of digital images. 4) \textbf{Caltech101} \cite{Caltech101} consists of 9,144 images following \cite{completer} with the views of HOG \cite{PHOG} and GIST features.
We conduct all the experiments on the platform of ubuntu 16.04 with Tesla P100 Graphics Processing Units (GPUs) and 32G memory size. Our model, method and baseline are built on the pytorch 1.7.0 framework.
\begin{table*}[ht!]
\caption{The performance comparison on multiple-view datasets. }
\label{table1}
\renewcommand\arraystretch{1.2}
\centering
\footnotesize
\setlength{\tabcolsep}{2.8mm}{
\begin{tabular}{lcccccccccccc}
\hline
Datasets               & \multicolumn{3}{c}{Sence-15}  & \multicolumn{3}{c}{Deep Animal} & \multicolumn{3}{c}{MNIST-USPS} & \multicolumn{3}{c}{Caltech101}\\
Evaluation metrics     & NMI    & ACC    & ARI        & NMI    & ACC    & ARI      & NMI    & ACC    & ARI     & NMI    & ACC    & ARI \\ \hline
COMPLETER \cite{completer}(2021)     & 0.382    & 0.147 & 0.218    & 0.387   & 0.274    & 0.161    & 0.523  & 0.216  & 0.341   &0.370 & 0.147  & 0.219\\
MvCLN \cite{MvCLN}(2021)    & 0.355    & 0.385    & 0.193    & 0.366    & 0.250    & 0.150   & 0.698  &0.810  & 0.590  &0.325  &0.175  &0.169 \\
SURE \cite{SURE}(2022)        & 0.319    & 0.392    & 0.195   & 0.339   & 0.251    & 0.159      & 0.653  & 0.830  & 0.661  &0.358  & 0.257  &0.234  \\
HmiMVC             (Ours)   & 0.395   & 0.398    & 0.221   & 0.398    & 0.236    & 0.198   & 0.726  & 0.782  & 0.658 &0.376  & 0.248  & 0.229\\
\hline
\end{tabular}
}
\end{table*}

Based on extensive ablation studies, the batch size is set to 1024 and the epochs of the three steps of training are 150,  respectively. We utilize Adam optimizer \cite{Adam} with default parameters and a learning rate of 0.0001. We set the dimension of the autoencoder and prediction model to d-1024-1024-1024-10-1024-1024-1024-d, where d is the dimension of the input data. We simply adopt a dense (i.e., fully-connected) network where each layer is followed by a batch normalization layer and a ReLU layer. The clustering effectiveness is evaluated by three metrics, i.e., normalized mutual information (NMI), clustering accuracy (ACC),  and adjusted Rand index (ARI).
 \begin{figure}[H]
\begin{minipage}[b]{.48\linewidth}
  \centering
  \centerline{\includegraphics[width=4.0cm]{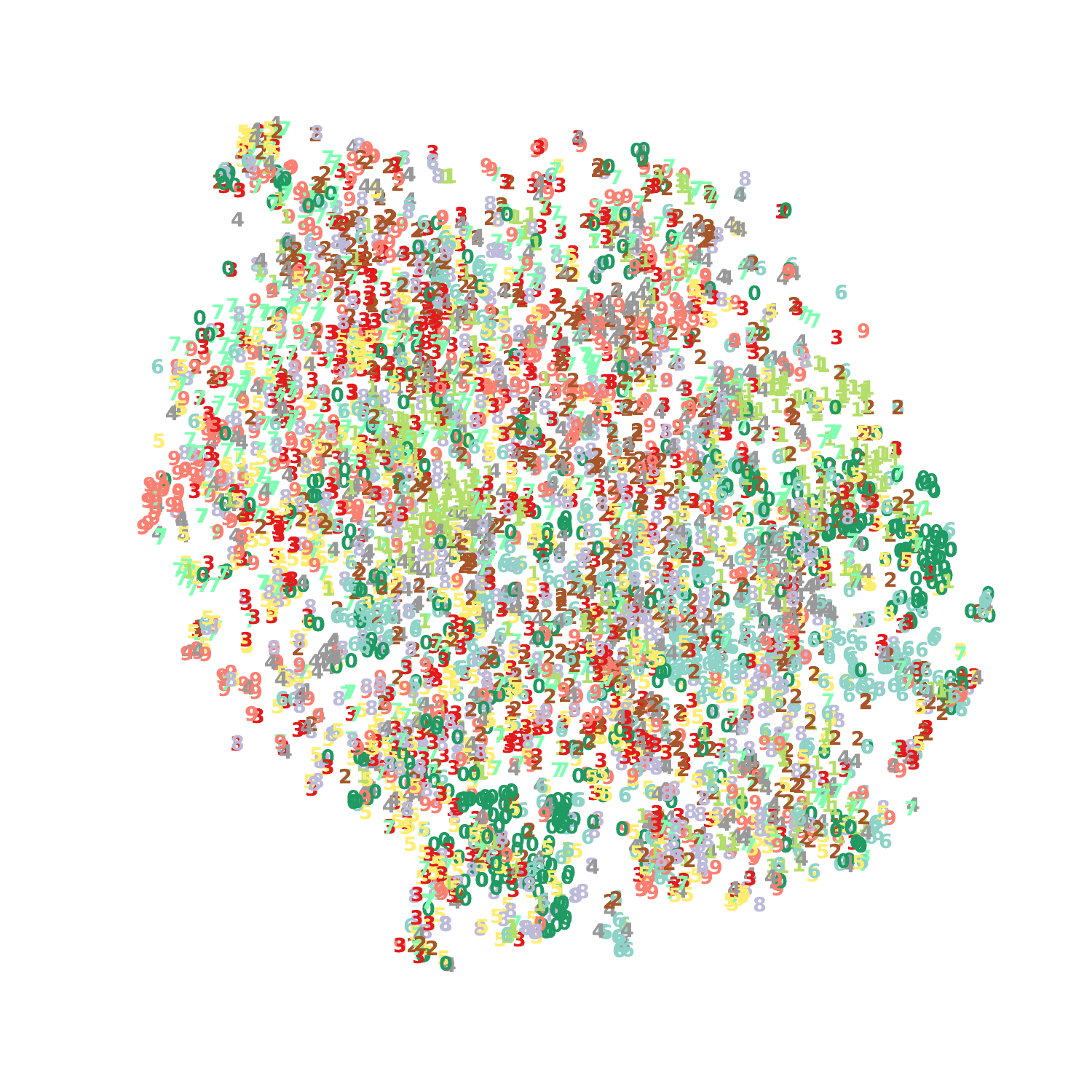}}
  \centerline{(a) Epoch 1 (NMI = 0.058)}\medskip
\end{minipage}
\hfill
\begin{minipage}[b]{0.48\linewidth}
  \centering
  \centerline{\includegraphics[width=4.0cm]{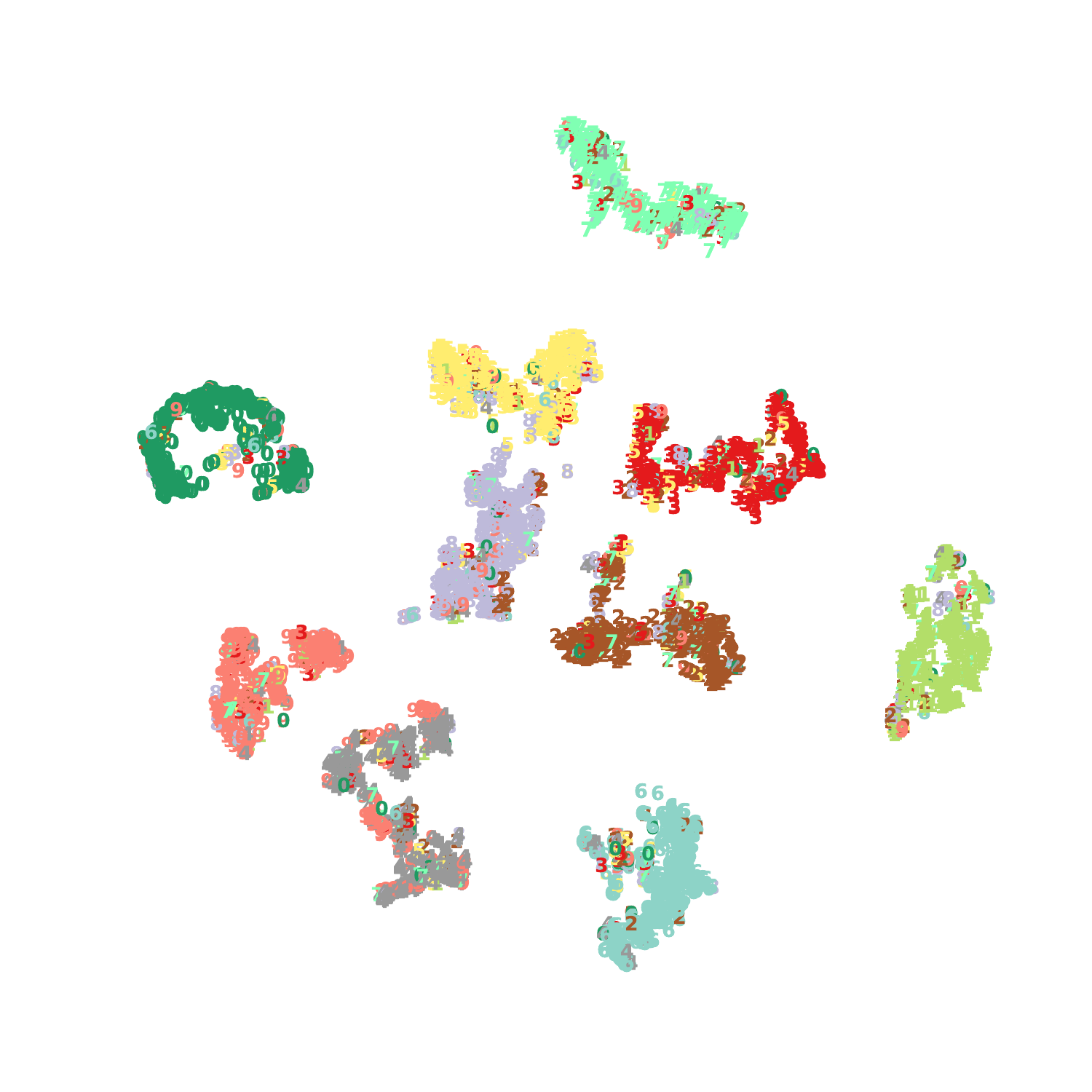}}
  \centerline{(b) Epoch 150 (NMI = 0.782)}\medskip
\end{minipage}
\caption{T-sne \cite{t-SNE} visualization on the MNIST-USPS dataset with increasing training iteration.}
\label{fig3}
\end{figure}

\begin{figure}[H]
\centering
\includegraphics[width=0.95\linewidth]{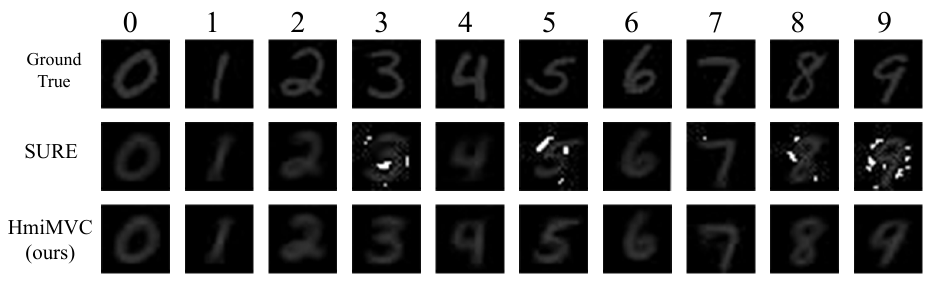}
\caption{Data recovery on noisy MNIST UPS datasets. Line 1 is the anchor view, while Line 2 and 3 are the SURE and HmiMVC view recovery results, respectively.}
\label{fig3.3}
\end{figure}
\subsection{Result Analysis}
In this section, to further verify the generalization of HmiMVC among different types of datasets, we conduct experiments in complex scenarios where both PVP and PSP coexist. To uniformly evaluate the performance of HmiMVC on incomplete unaligned multi-view data,  we randomly select ${N}_{x}$, ${N}_{w}$, and ${N}_{s}$ instances as training data, incomplete data, and unaligned data, respectively. We compare HmiMVC with 3 multi-view clustering baselines, including COMPLETER \cite{completer}, MvCLN \cite{MvCLN}, SURE. For all methods, we use the recommended network structure and parameters for fair comparison, with the complete data ratio is set to $\gamma=0.5$. The ratio of alignment data $\beta$ and missing data $\alpha$ is set to 0.25 to simulate PVP and PSP. Since COMPLETER cannot handle unaligned data, we establish the alignment relationship directly with Hungarian algorithm. And SURE and MvCLN can only be aligned on complete data, so 50\% of the data is filled first before alignment, and in MvCLN we fill with the mean value of the same view.
Table \ref{table1} depicts the clustering performances of all methods, our HmiMVC achieves the best NMI on four datasets, the best ACC on Sence-15 and the best ARI on Deep Animal and MNIST-USPS. 
ACC and ARI determine whether the cluster result is similar to the ground-true. With advanced hierarchical information learning (the best NMI values), the clustering results become more compact and independent through training, and achieve more accurate clustering.  
The visualization of the learned features (Fig.\ref{fig3}) also illustrates the same trend. Compared to SURE, HimMVC recovers less view noise (Fig.\ref{fig3.3}).

\subsection{Model Analysis}

\subsubsection{Convergence Analysis}
\begin{figure}[H]
\centering
\includegraphics[width=0.7\linewidth]{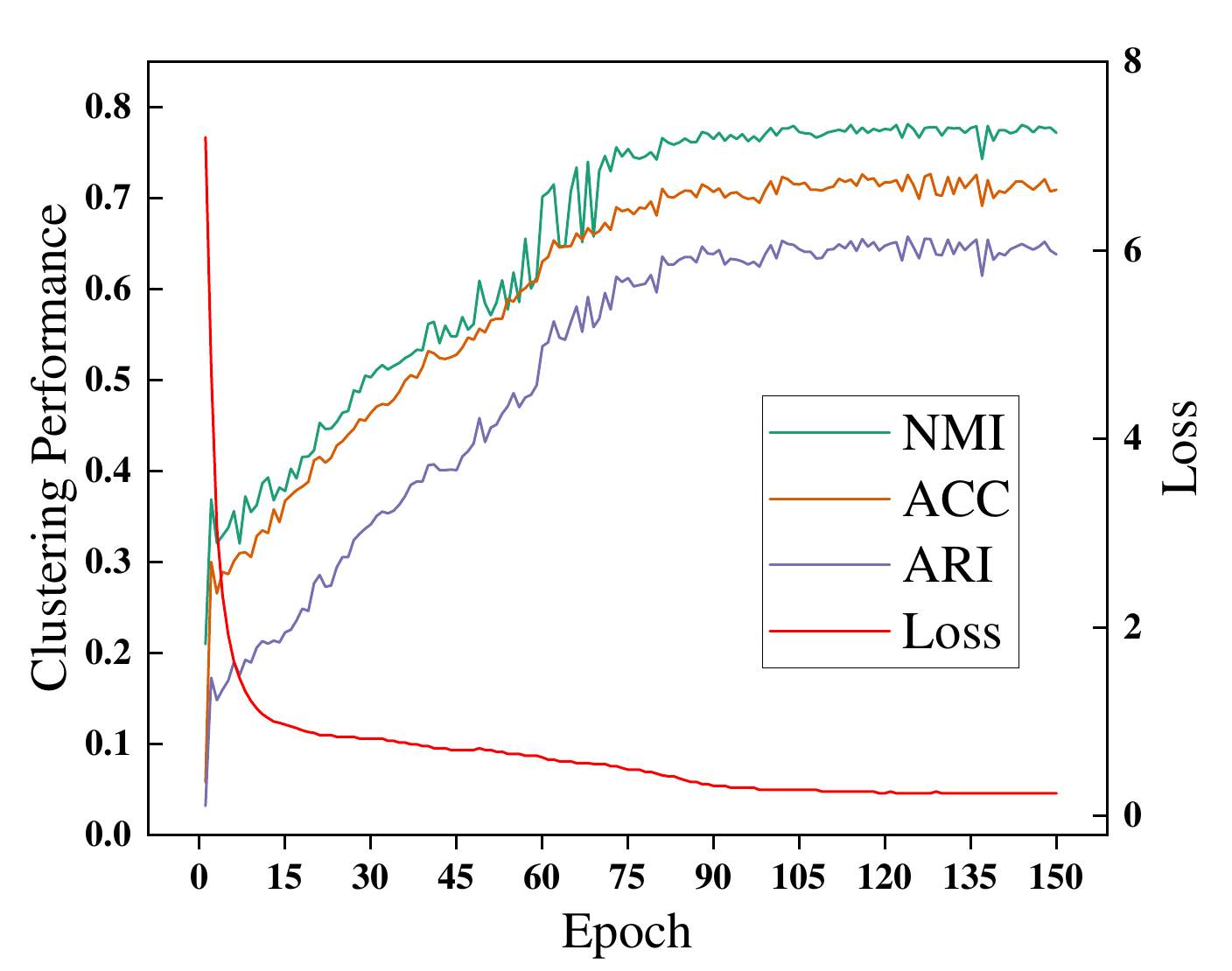}
\caption{Convergence analysis of clustering performance and loss values.}
\label{fig4.1}
\end{figure}

We investigate the convergence of HmiMVC by reporting the loss value and the corresponding clustering performance with increasing epochs. As shown in Fig. \ref{fig4.1}, one could observe that the loss remarkably decreases in the first 20 epochs, and various evaluation metrics continuously increases and tends to be smooth and consistent.

\subsubsection{Parametric Analysis}
\begin{figure}[H]
\centering
\includegraphics[width=0.6\linewidth]{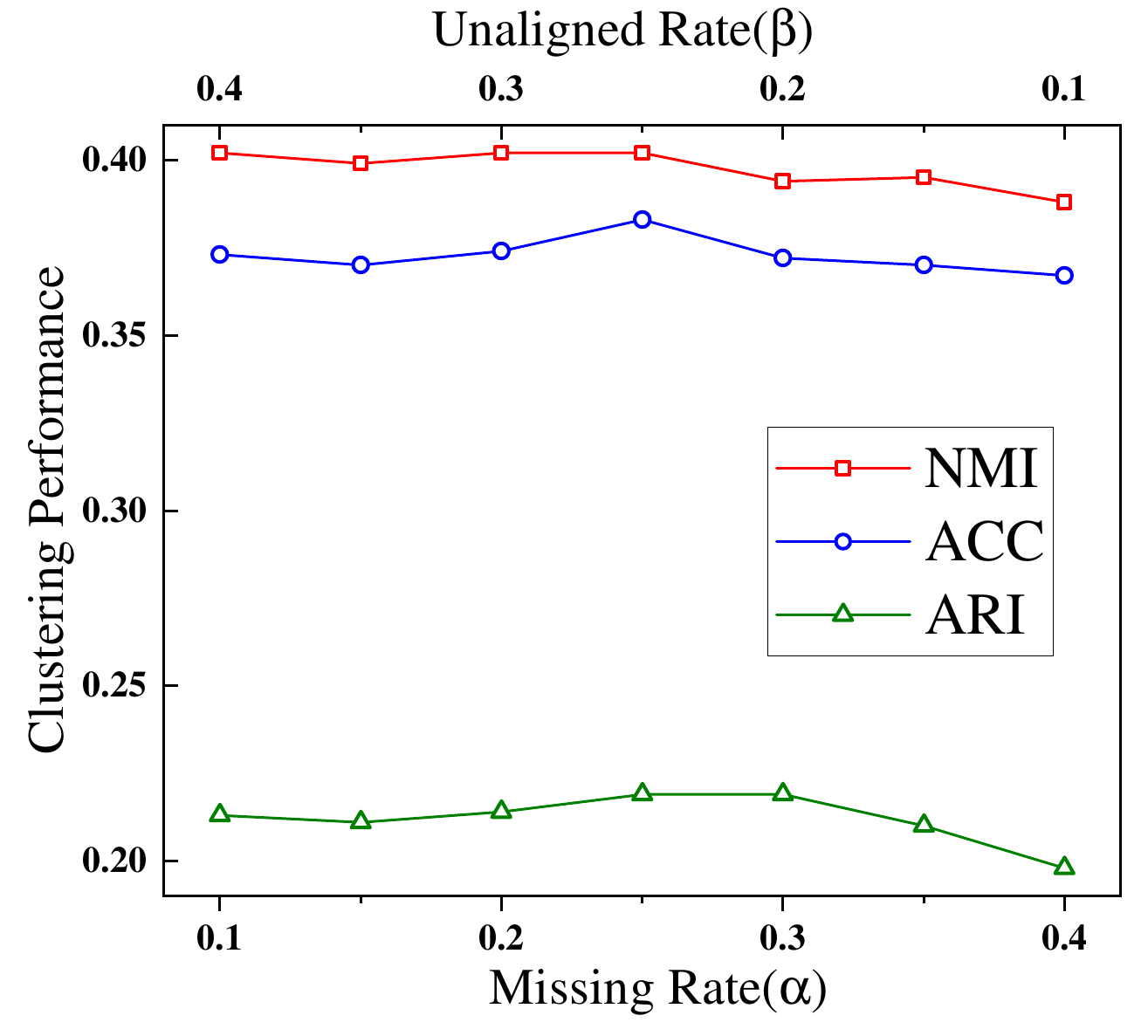}
\caption{Parameter analysis of performance comparisons with different missing rates ($\alpha$) and unaligned rates ($\beta$).}
\label{fig4.2}
\end{figure}

As illustrated in the Fig.\ref{fig4.2}, with the complete rate $\gamma$, our clustering performance is always relatively stable regardless of different $\beta$ and $\alpha$. This demonstrates that HmiMVC has strong robustness.

\subsubsection{Ablation experiment}
We conduct the ablation study on MNIST-USPS to demonstrate the importance of each component of our method. As shown in Table \ref{table2}, all losses play an integral role in HmiMVC. Only when the alignment strategy of the second level, $\mathcal{L}_ {cl}$, is involved, a significant improvement of clustering performance can be observed. 
\begin{table}[htbp]
\renewcommand\arraystretch{1.3}
\centering
\footnotesize
\caption{Ablation study}
\label{table2}
\setlength{\tabcolsep}{2.5mm}{
\begin{tabular}{lccc}
\text moudules of HmiMVC                          & NMI     & ACC    & ARI     \\
\hline \text (1)$\mathcal{L}_{pre}$         & 0.169   & 0.211  & 0.092  \\
             (2)$\mathcal{L}_{rec}$         & 0.329   & 0.457  & 0.242  \\
             (3)$\mathcal{L}_{cl}$          & 0.485   & 0.541 & 0.234  \\
             (4)$\mathcal{L}_{rec}+\mathcal{L}_ {pre}$        & 0.486   & 0.523  & 0.347  \\
             (5)$\mathcal{L}_{pre}+\mathcal{L}_ {cl}$         & 0.539   & 0.607  & 0.413  \\
             (6)$\mathcal{L}_{rec}+\mathcal{L}_{cl}$           & 0.693   & 0.760  & 0.675  \\
             (7)$\mathcal{L}_{rec}+\mathcal{L}_{pre}+\mathcal{L}_{cl}$         & 0.726  & 0.782  & 0.658  \\
\hline

\end{tabular}
}
\end{table}

\section{CONCLUSION}
\label{CONCLUSION}

This paper proposes HmiMVC to provide a hierarchically consistent framework for handling PVP along with PSP. 
HmiMVC achieves the consistency learning across views by maximizing the hierarchical mutual information and minimizing the conditional entropy, bridging the  gap of existing schemes.
To the best of our knowledge, this is the first work that class-level and instance-level alignment strategies have been combined, which has enabled HmiMVC to achieve state-of-the-art performance in practice.
We experimentally show that our loss could mitigate or even eliminates the noise introduced during pairwise construction.
This framework trains feature extractors and predictors, which can be used in areas such as feature compression, unsupervised labeling, and cross-modal feature retrieval. 
In the future, we plan to further extend to work in the presence of a higher number of views.




\begin{IEEEbiography}
[{\includegraphics[width=1in,height=1.25in,clip,keepaspectratio]{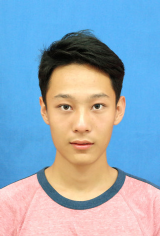}}]{Jiatai Wang}{\space}received M.S. degree from Inner Mongolia University of Technology, Hohhot, China. Recently, he is working as a visiting scholar and going to pursue his Ph.D degree at Institute of Computing Technology, Chinese Academy of Sciences, Beijing, China. He has published several papers in high impact journals in computer vision field, such as IET computer vision. His interests are focused on unsupervised learning in CV field. 
\end{IEEEbiography}
\begin{IEEEbiography}
[{\includegraphics[width=1in,height=1.25in,clip,keepaspectratio]{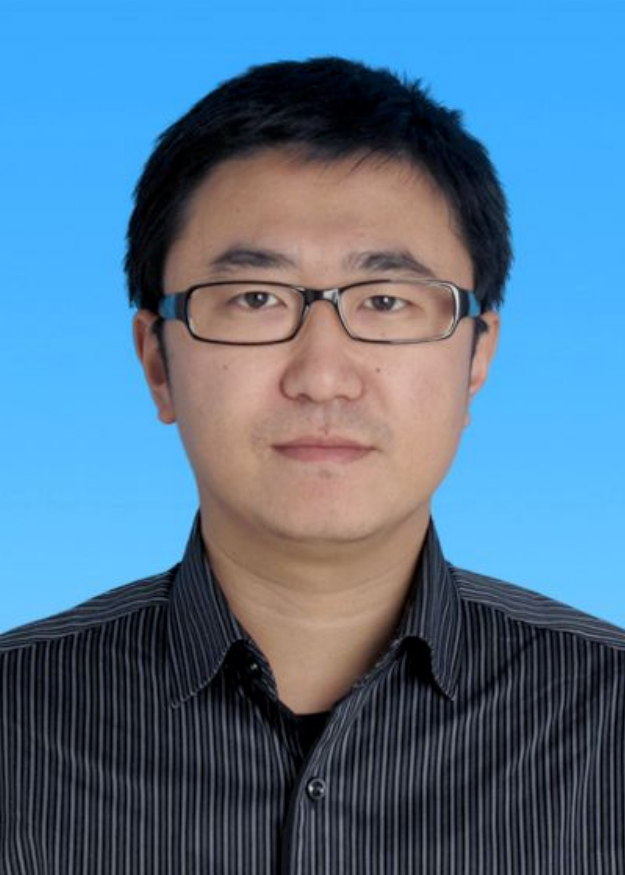}}]{Zhiwei Xu}{\space}received the B.S. degree in 2002 from University of Electronic Science and Technology of China, Chengdu, China, and the Ph.D. degree in 2018 from Institute of Computing Technology, Chinese Academy of Sciences, Beijing, China. He is an associate professor and M.S. supervisor of Inner Mongolia University of Technology, while working as an Adjunct Professor in Institute of computing, Chinese Academy of Sciences. From 2020 to 2021, he worked towards visiting post-doctoral in the Department of Electrical and Computer Engineering, State University of New York at Stony Brook, Stony Brook, NY. His research interests include in-network data compact representation, learning, and the related security and privacy problems.
\end{IEEEbiography}
\begin{IEEEbiography}
[{\includegraphics[width=1in,height=1.25in,clip,keepaspectratio]{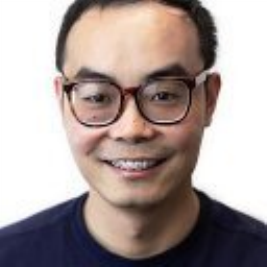}}]{Xuewen Yang}{\space}, an accomplished scholar specializing in Multimodal Learning and Natural Language Processing, is currently a Senior Researcher at InnoPeak Technology, Palo Alto, US. He earned his B.S. and M.S. from Xi'an Jiaotong University, China, and another M.S. from Ecole Centrale Marseille, France. He completed his Ph.D. at Stony Brook University, New York, in 2021. With a focus on bridging theory and practice, Dr. Yang is driving technological innovation in computational systems through his research.
\end{IEEEbiography}
\begin{IEEEbiography}
[{\includegraphics[width=1in,height=1.25in,clip,keepaspectratio]{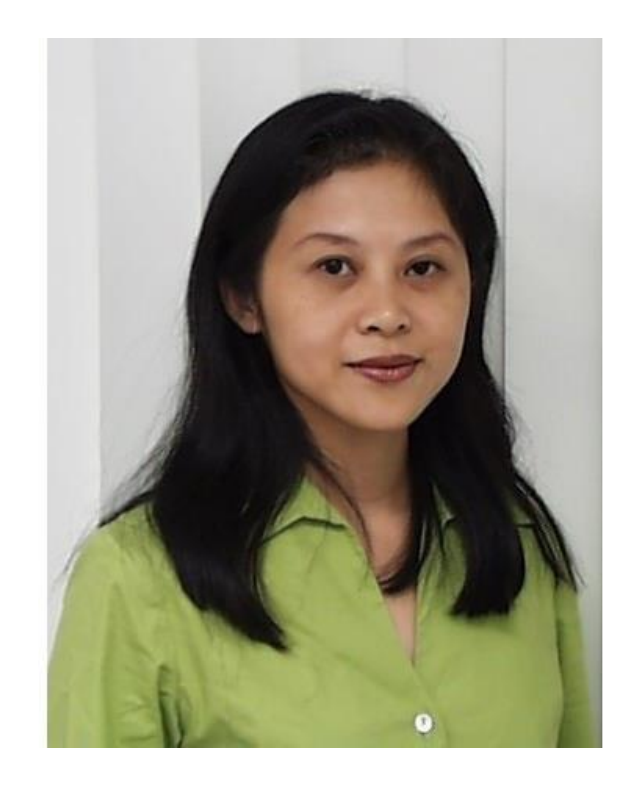}}]{Xin Wang}{\space}received the B.S. and M.S. degrees in telecommunications
engineering and wireless communications engineering respectively from Beijing University of Posts and Telecommunications, Beijing, China, and the Ph.D. degree in electrical and computer engineering from Columbia University, New York, NY. She is currently an Associate Professor in the Department of Electrical and Computer Engineering of the State University of New York at Stony Brook, Stony Brook, NY. Before joining Stony Brook, she was a Member of Technical Staff in the area of mobile and wireless networking at Bell Labs Research, Lucent Technologies, New Jersey, and an Assistant Professor in the Department of Computer Science and Engineering of the State University of New York at Buffalo, Buffalo, NY. Her research interests include algorithm and protocol design in wireless networks and communications, mobile and distributed computing, as well as big data analysis and machine learning. She has served in executive committee and technical committee of numerous conferences and funding review panels, and serves as the associate editor of IEEE Transactions on Mobile Computing. Dr. Wang achieved the NSF career award in 2005, and ONR challenge award in 2010.
\end{IEEEbiography}
\end{document}